\documentclass[3p,authoryear]{elsarticle}

\usepackage{amsmath}
\usepackage{amssymb}
\usepackage{bbm}
\usepackage{bm}

\usepackage{afterpage}
\usepackage{pdflscape}
\usepackage{hyperref}

\journal{Computational Statistics and Data Analysis}

\begin{document}

\begin{frontmatter}

\title{LogitBoost autoregressive networks}

\author{Marc Goessling\corref{mycorrespondingauthor}}
\address{Department of Statistics, University of Chicago,
USA}
\cortext[mycorrespondingauthor]{Corresponding author.}
\ead{goessling@uchicago.edu}

\begin{abstract}
Multivariate binary distributions can be decomposed into products of univariate conditional distributions. Recently popular approaches have modeled these conditionals through neural networks with sophisticated weight-sharing structures. It is shown that state-of-the-art performance on several standard benchmark datasets can actually be achieved by training separate probability estimators for each dimension. In that case, model training can be trivially parallelized over data dimensions. On the other hand, complexity control has to be performed for each learned conditional distribution. Three possible methods are considered and experimentally compared. The estimator that is employed for each conditional is LogitBoost. Similarities and differences between the proposed approach and autoregressive models based on neural networks are discussed in detail.
\end{abstract}

\begin{keyword}
Autoregressive model \sep Conditional distribution \sep Boosting
\end{keyword}

\end{frontmatter}

\section{Introduction}
Given a collection of multivariate binary data points $\bm{x}^{(n)} \in \{0,1\}^D$, $n=1,\ldots,N$, we consider the fundamental task of learning the underlying distribution $\mathbb{P}(\bm{x})$. Distribution estimates have many potential applications. For example, they can be used for anomaly detection, denoising, data synthesis, missing value imputation, classification and compression. A useful fact is that every multivariate distribution can be factored into a product of univariate conditional distributions
\begin{equation}
\mathbb{P}(\bm{x}) = \prod_{d=1}^D \mathbb{P}(x_d \,|\, \bm{x}_{1:d-1}).
\label{eq:joint_decomposition}
\end{equation}
Thus, it is always possible to learn a joint distribution by estimating univariate conditionals. The decomposition (\ref{eq:joint_decomposition}) is known as a fully connected left-to-right graphical model or autoregressive network \citep{frey1998graphical}. Fitting a sequence of univariate conditional distributions can be easier than trying to model high-dimensional dependencies directly. Moreover, with the factorization it is straightforward to perform likelihood evaluations and exact sampling.

Many different approaches have been proposed to model the conditional distributions in an autoregressive network. In the simplest approach \citep{cooper1992bayesian} the conditional distributions are learned by performing a greedy search for a sparse set $\mathrm{par}(d) \subset \{1,\ldots,d-1\}$ of predictors for each dimension $d$. The conditional probabilities $P(x_d \,|\, \bm{x}_{\mathrm{par}(d)})$ are then simply estimated through the corresponding empirical frequencies. However, since the number of parameters that have to be learned grows exponentially in the number of predictors, only very few predictors can be used.

A fruitful alternative is to approximate the conditional distributions through parametric models. For example, \citet{frey1998graphical} considered logistic autoregressive networks, which are also called fully visible sigmoid belief nets. In these networks the log-odds of $x_d=1$ are modeled as a linear function of $\bm{x}_{1:d-1}$. That means
\[
\mathbb{P}(x_d{=}1 \,|\, \bm{x}_{1:d-1}) = \sigma(\alpha^{(d)}_1 x_1 + \ldots + \alpha^{(d)}_{d-1} x_{d-1}),
\]
where $\bm{\alpha}^{(d)} \in \mathbb{R}^{d-1}$ and $\sigma(z) = (1+\exp(-z))^{-1}$ is the logistic function. However, the performance of these simple networks can be limited if the true conditional distributions are highly nonlinear. Mixtures of such networks \citep{goessling2016mixtures} often lead to improvements if sufficient data is available.

More ambitious approaches attempt to directly model the log-odds of $x_d=1$ as a nonlinear function of $\bm{x}_{1:d-1}$. One major research direction focuses on neural networks for this task. These models are of the form
\begin{equation}
\mathbb{P}(x_d{=}1 \,|\, \bm{x}_{1:d-1}) = \sigma(b^{(d)} + (\bm{v}^{(d)})^T\bm{h}^{(d)}),
\label{eq:neural_conditional}
\end{equation}
where $b^{(d)} \in \mathbb{R}$, $\bm{v}^{(d)} \in \mathbb{R}^H$. In the original work by \citet{bengio2000taking}, each data dimension was modeled through a different set of (deterministic) hidden units
\[
\bm{h}^{(d)} = \sigma(\bm{c}^{(d)}+\bm{W}^{(d)}\bm{x}_{1:d-1}) \in \mathbb{R}^H,
\]
where $\bm{c}^{(d)} \in \mathbb{R}^H$, $\bm{W}^{(d)} \in \mathbb{R}^{H \times (d-1)}$. However, that approach suffered from substantial overfitting. To improve the generalization performance it was suggested to prune connections after the training phase, based on statistical tests. In other words, some entries in $\bm{W}^{(d)}$ were manually set to 0. \citet{larochelle2011neural} subsequently introduced the idea of weight sharing between the parameters for different dimensions.
Specifically, they proposed to use the same bias term $\bm{c}^{(d)} = \bm{c}$ for all dimensions $d$ and connections of the form $\bm{W}^{(d)} = \bm{W}_{\cdot,1:d-1}$, where $\bm{W} \in \mathbb{R}^{H\times(D-1)}$. This approach led to drastically better results \citep{bengio2011discussion}. Many variations of their idea \citep[e.g.][]{uria2014deep,raiko2014iterative,germain2015made} have been considered afterwards. The main difference between the proposed models is the number of hidden layers, the connections within and between layers, and the choice of the activation function. The more advanced models use an ensemble of networks corresponding to different orderings of the data dimensions or corresponding to different connectivity patterns. An equal mixture of these networks is then used as the final model.

A rather different class of nonparametric estimators are decision trees \citep{breiman1984classification}. When employed for probabilistic modeling they are also known as probability estimation trees \citep{provost2003tree}. These trees model conditional probabilities through piecewise-constant functions
\[
\mathbb{P}(x_d{=}1 \,|\, \bm{x}_{1:d-1}) = \sum_{j=1}^J p_j^{(d)} \mathbbm{1}(\bm{x}_{1:d-1}{\in}R_j^{(d)}),
\]
where the regions $R_j^{(d)}$ partition the predictor space $\{0,1\}^{d-1}$ and $p_j^{(d)} \in (0,1)$. Since the predictors are binary, each region is defined by specifying the value of some of the variables $x_1,\ldots,x_{d-1}$. Decision trees as models for conditional distributions have been used for example by \citet{boutilier1996context,friedman1998learning}. Decision trees are attractive because they are flexible, interpretable and computationally efficient. The performance of single trees is often unsatisfactory though because shallow trees can have a large bias and deep trees often have a large variance.

However, decision trees can be assembled into a powerful estimator through a method called boosting \citep{friedman2001greedy}. This procedure creates a collection of relatively shallow decision trees, which are trained as to complement each other. The bias-variance tradeoff for boosted trees is often much better than for single decision trees. Particularly relevant for us is LogitBoost \citep{friedman2000additive}, which is a probability estimator based on boosted trees.

In this work, we explore the use of LogitBoost for learning high-dimensional binary distributions from data. We call our model a LogitBoost autoregressive network (LBARN). Our contributions are as follows:
\begin{enumerate}
\item We propose LogitBoost as a learning procedure for the conditionals in an autoregressive network. In existing work \citep{shafik2009boosting} boosting was applied to continuous, stationary time series. Thus, only a single conditional distribution had to be learned. Our network, on the other hand, can train several hundred conditional distributions (one for each data dimension).
\item Neural autoregressive networks are currently the state-of-the-art for nonparametric distribution estimation in high-dimensions (say $D>50$). We show that there is a simpler alternative that works equally well. Similarities and differences between our approach and neural methods are discussed in detail.
\item In contrast to (most) neural autoregressive networks our model does not use any weight sharing among conditionals for different dimensions. Consequently, our training procedure can be perfectly parallelized over the data dimensions. However, this also means that we need a method to perform complexity control for each conditional distribution. We propose three different procedures and empirically evaluate them on several diverse datasets.
\item Finally, we study how the ordering of the variables affects the performance of the trained model. In particular, we introduce a sorting procedure that allows us to build a simple compression mechanism.
\end{enumerate}

We start in Section \ref{sec:logitboost} by providing an overview of LogitBoost. In Section \ref{sec:model_selection} we then describe how separate LogitBoost runs for each dimension can be combined into a coherent joint model. In Section \ref{sec:neural_comparison} we contrast our model with neural autoregressive networks. In Section \ref{sec:experiments} we present quantitative results for several common datasets and discuss the choice of hyperparameters. We also show how the strength of (nonlinear) dependencies between the variables can be quantified in our model and illustrate that the conditional distributions can be learned robustly even when a large number of irrelevant variables are present. Moreover, we consider how the autoregressive network is affected when different learned orderings of the variables are used.

\section{LogitBoost}
\label{sec:logitboost}
LogitBoost \citep{friedman2000additive} is a forward stagewise procedure that fits additive logistic regression models\footnote{The model $\mathbb{P}_0$ is defined as a Bernoulli-1/2 distribution, independently of $\bm{x}$.}
\[
\mathbb{P}_t(y{=}1 \,|\, \bm{x}) = \sigma(f_1(\bm{x})+\ldots+f_t(\bm{x}))
\]
for $t=1,\ldots,T$ to the training data $(\bm{x}^{(n)},y^{(n)}) \in \{0,1\}^D \times \{0,1\}$, $n=1,\ldots,N$, by maximum likelihood. While in the context of generalized additive models \citep{hastie1990generalized} additivity refers to a sum of (smooth) univariate functions, here additivity is meant in the sense of a sum of multivariate functions. The optimization is based on a second-order Taylor expansion of the Bernoulli log-likelihood
\[
L_t = \sum_{n=1}^N \log \mathbb{P}_t(y^{(n)} \,|\, \bm{x}^{(n)}).
\]
\citet{friedman2000additive} derived that
\[
\frac{\partial L_t}{\partial f_t} = \sum_{n=1}^N (y^{(n)} - p_{t-1}^{(n)}), \quad \frac{\partial^2 L_t}{\partial f_t^2} = -\sum_{n=1}^N p_{t-1}^{(n)}(1-p_{t-1}^{(n)}),
\]
where 
\[
p_{t-1}^{(n)} = \mathbb{P}_{t-1}(y{=}1 \,|\, \bm{x}^{(n)})
\]
is the conditional success probability given $\bm{x}^{(n)}$ as predicted by the model after $t-1$ rounds of boosting. The Newton step results in a weighted least-squares regression of the current pseudoresiduals $y^{(n)} - p_{t-1}^{(n)}$ onto the predictors $\bm{x}^{(n)}$. The base learners are typically $J$-terminal regression trees
\[
f_t(\bm{x}) = \sum_{j=1}^J \gamma_j \mathbbm{1}\{\bm{x}{\in}R_j\},
\]
where $\gamma_j \in \mathbb{R}$. It can be shown \citep{li2010robust} that the weighted regression problem is then equivalent to fitting a decision tree with a simple objective function. Specifically, the objective value for a subset $R$ of the predictor space $\{0,1\}^D$ is
\[
\mathrm{obj}(R) = \frac{\left[\sum\limits_{n=1}^N (y^{(n)}-p_{t-1}^{(n)}) \mathbbm{1}(\bm{x}^{(n)}{\in}R)\right]^2}{\sum\limits_{n=1}^N p_{t-1}^{(n)}(1-p_{t-1}^{(n)}) \mathbbm{1}(\bm{x}^{(n)}{\in}R)}.
\]
The task is to choose the regions $R_j \subset \{0,1\}^D$, $j=1,\ldots,J$, in such a way that $\sum_{j=1}^J \mathrm{obj}(R_j)$ is maximized. Since the exact optimization is very expensive, this is usually done approximately through a greedy procedure \citep{breiman1984classification}. One starts with a single region, corresponding to the entire predictor space $\{0,1\}^D$, and then recursively divides the region that leads to the largest improvement of the objective. The gain of splitting a region $R$ using variable $x_d$ is
\begin{equation}
\mathrm{gain}(d) = \mathrm{obj}(\{\bm{x} \in R \,|\, x_d{=}0\}) + \mathrm{obj}(\{\bm{x} \in R \,|\, x_d{=}1\}) - \mathrm{obj}(R).
\label{eq:gain}
\end{equation}
Once a $J$-terminal tree is learned the value of the leaves is set to
\begin{equation}
\gamma_j = \frac{\sum\limits_{n=1}^N (y^{(n)}-p_{t-1}^{(n)}) \mathbbm{1}(\bm{x}^{(n)}{\in}R_j)}{\sum\limits_{n=1}^N p_{t-1}^{(n)}(1-p_{t-1}^{(n)}) \mathbbm{1}(\bm{x}^{(n)}{\in}R_j)}.
\label{eq:leaf_update}
\end{equation}
The number of leaves $J$ impacts the depth of the trees and hence determines the order of interactions that can be modeled. While LogitBoost is rather robust to overfitting for classification tasks, overfitting does occur for probability estimates if too many rounds of boosting are performed \citep{mease2007boosted}. It is therefore customary to choose the number of trees based on the performance on holdout data. Better results can often be achieved by introducing shrinkage since this facilitates the use of somewhat deeper trees, which otherwise would not generalize well. The log-odds of $y=1$ in this case is modeled as $\nu \sum_{t=1}^T f_t$, where $\nu \in (0,1)$ is a shrinkage parameter. Smaller values of $\nu$ almost always lead to better generalization performance. However, if $\nu$ is very small then many trees are needed and hence the computational costs grow accordingly.

\section{Model selection}
\label{sec:model_selection}
Our main proposal is to apply LogitBoost separately for each data dimension $d=1,\ldots,D$ and to learn the conditional distributions $\mathbb{P}(x_d \,|\, \bm{x}_{1:d-1})$ using up to $T$ boosted trees. This yields $D$ sequences of models $\mathbb{P}_t(x_d \,|\, \bm{x}_{1:d-1})$, $t=0,\ldots,T$, with increasing complexity. In order to obtain a single joint model $\mathbb{P}(\bm{x})$ for the full data vector, we have to choose the number of boosting rounds $t_d \in \{0,\ldots,T\}$ for each dimension $d$. Thus, in total there are $(T+1)^D$ candidate models for the joint distribution. In the following we propose three different strategies for selecting one of those candidates.

\subsection{Individual selection}
A straightforward approach is to perform model validation individually for each dimension $d$. This means we compute the likelihood of holdout data $x_d$ given $\bm{x}_{1:d-1}$ under each of the models $\mathbb{P}_t(x_d \,|\, \bm{x}_{1:d-1})$, $t=0,\ldots,T$, and find the number of boosting rounds $t_d^\star$ that corresponds to the model with the highest validation likelihood for the $d$-th dimension. The joint distribution is then estimated through
\[
\mathbb{P}_\textrm{ind}(\bm{x}) = \prod_{d=1}^D \mathbb{P}_{t_d^\star}(x_d \,|\, \bm{x}_{1:d-1}).
\]
If the validation set is rather small then this approach can actually be expected to overfit the validation data since we are making $D$ model choices. The next two approaches, on the other hand, perform only a single model selection.

\subsection{Common selection}
A conservative method for model selection is to use the entire data vector $\bm{x}$ for validation and to choose a common number of trees for all dimensions. In this case we compute the validation likelihood for the models $\prod_{d=1}^D \mathbb{P}_t(x_d \,|\, \bm{x}_{1:d-1})$, $t=0,\ldots,T$, and find the maximizer $t^\star$. The joint distribution is then estimated through
\[
\mathbb{P}_\textrm{com}(\bm{x}) = \prod_{d=1}^D \mathbb{P}_{t^\star}(x_d \,|\, \bm{x}_{1:d-1}).
\]
This approach implicitly assumes that all conditional distributions have about the same complexity. Hence, if the individual variables are heterogeneous (in terms of variance) then the common selection of boosting rounds may hurt the performance of the distribution estimator.

\subsection{Linearized selection}
\label{sec:linearized_selection}
A third approach for model selection is to construct a linear ordering of all learned trees and to perform validation on the corresponding sequence of joint models. Specifically, we can start with ``empty'' models $\mathbb{P}_0(x_d \,|\, \bm{x}_{1:d-1})$ for all dimensions, i.e., using zero trees. We then greedily add a tree to the dimension that gives rise to the largest improvement in terms of training likelihood for the full data vector $\bm{x}$. Alternatively, the sorting procedure can be performed in a backward manner by starting with the ``full'' model $\mathbb{P}_T(x_d \,|\, \bm{x}_{1:d-1})$ for each dimension, i.e., using all $T$ trees. We then greedily remove the tree that gives rise to the smallest decrease of training likelihood. Both procedures create a sequence of models $\mathbb{P}_s(\bm{x})$, $s=0,\ldots,D \cdot T$.
The joint distribution can then be estimated through
\[
\mathbb{P}_\textrm{lin}(\bm{x}) = \mathbb{P}_{s^\star}(\bm{x}),
\]
where $s^\star$ is the maximizer of the likelihood for holdout data $\bm{x}$. The linearized selection method makes only a single model selection but allows for different model complexities of the individual dimensions.

\section{Comparison with neural autoregressive networks}
\label{sec:neural_comparison}

Our model as well as autoregressive models based on neural networks learn a nonlinear function of $\bm{x}_{1:d-1}$ for the (conditional) log-odds of $x_d=1$. We continue with a discussion of similarities and differences between these two approaches.

\subsection{Parallelization}
\label{sec:parallelization}
One major difference is that we estimate separate conditional models for each dimension. This means that our training procedure can be perfectly parallelized over the data dimensions. Indeed, each conditional distribution can be learned on a separate processor without requiring any communication. The fact that state-of-the-art performance can be achieved without any weight sharing (as we will show in Section \ref{sec:experiments}) is a nontrivial result. The parameter sharing introduced by \citet{larochelle2011neural} greatly improved the generalization performance of neural autoregressive networks and there are no reported competitive results for neural approaches without extensive parameter sharing of this type.

\subsection{Architecture}
Another fundamental difference is that neural networks have a fixed architecture, which is chosen by trial and error. This includes the number of hidden layers, the number of units per layer and the connections between them (including the choice of activation function). Only the weights for the connections are learned. In contrast to that, LogitBoost also learns the structure of the nonlinear functions. Indeed, each decision tree produces a function that depends only on a small subset of the variables. The number of tree leaves $J$ determines what degree of interactions between the predictors can be incorporated. Such a hyperparameter does not exist in neural networks.

\subsection{Optimization}
A further distinction between the two approaches lies in the optimization procedure. Neural networks use backpropagation, which refers to gradient descent on the log-likelihood of the model. The learning process is started from a random initialization of the parameters. Our model, on the other hand, starts with log-odds of zero for each conditional and uses a second-order Taylor approximation of the log-likelihood for the parameter updates. In both cases the model complexity increases with additional iterations of gradient descent or additional rounds of boosting. The learning rate for neural networks has a similar role as the shrinkage factor $\nu$ for boosting.

Model selection for neural networks is performed by keeping track of the validation likelihood over iterations of gradient descent. The model with the best performance on the validation set is returned. Since all conditionals are learned at once only a single model selection is needed. The most analogous selection procedure in our approach is the one based on a linear ordering of the trees for all dimensions (see Section \ref{sec:linearized_selection}).

\subsection{Likelihood evaluation}
\label{sec:likelihood_evaluation}
A comprehensive list of the computational complexities for likelihood evaluation in neural autoregressive networks was provided by \citet{germain2015made}. The simplest autoregressive model based on neural networks is NADE \citep{larochelle2011neural}, which consists of a single hidden layer. Likelihood evaluations for NADE can be performed with $\mathcal{O}(DH)$ operations, where $H$ is the number of hidden units. Deeper versions of NADE \citep{uria2014deep} with multiple layers of hidden units require $\mathcal{O}(DH^2)$ operations though. Autoencoder networks \citep{germain2015made} with more than one layer of hidden units require $\mathcal{O}(DH+H^2+D^2)$ operations. For ensemble networks \citep{uria2014deep} the complexities scale multiplicatively with the number of different orderings of the data dimensions. Neural networks, which use stochastic hidden units \citep{gregor2014deep,bornschein2015reweighted}, require calculations that are exponential in $H$ because a sum over all possible hidden configurations has to be computed. In that case approximations via importance sampling are typically used \citep{salakhutdinov2008quantitative}. In contrast to that, the computational complexity to evaluate the exact log-likelihood of a test sample under our model is $\mathcal{O}(DTL)$, where $L$ is the average tree depth. If the trees are balanced then $L = \log J$. Thus, likelihood evaluations with our approach are orders of magnitude faster compared to the sophisticated variants of neural autoregressive networks.

\subsection{Potential benefits of our approach}
We summarize what we believe are benefits of our approach over neural autoregressive networks. As discussed in Section \ref{sec:parallelization} and \ref{sec:likelihood_evaluation}, our model has computational advantages because the training procedure can be parallelized and likelihoods can be calculated efficiently. This makes it possible to scale the approach to very high dimensions.

We see an additional advantage in terms of hyperparameter tuning. The only two hyperparameters that we have to choose in our approach are the shrinkage factor $\nu$ and the number of tree leaves $J$ (the number of trees $T$ is chosen through one of the model selection procedures from Section \ref{sec:model_selection}). Choosing $\nu$ is easy. Indeed, with two exceptions (for speed reasons) we were able to use the same shrinkage factor for all our experiments (Section \ref{sec:experiments}). To choose $J$ we explored a handful of possible values. In contrast to that, successfully tuning neural networks is still considered an art \citep{bengio2013representation}. Besides the (initial) learning rate, the number of gradient descent iterations and the architecture of the neural network there are several ``optimization tricks'' that might be required in order to achieve satisfactory performance. Typical factors, which affect the convergence, are the mini-batch size, the learning rate decay schedule and the momentum of the gradient descent learning algorithm. Moreover, careful initialization is important and additional regularization procedures like weight decay or dropout \citep{srivastava2014dropout} might be needed to generalize well.

The capacity of neural networks is typically chosen such that all potentially useful connections exist. Indeed, the functional structure (\ref{eq:neural_conditional}) of the conditional distributions in neural autoregressive networks implies that $x_d$ always depends on all predictors $\bm{x}_{1:d-1}$. The hope is that at the time of convergence the connections to irrelevant variables have a weight close to zero.
In contrast to that, our model automatically selects relevant variables because LogitBoost builds the functional structure of the conditional distributions in a sequential manner. Apart from being more interpretable, sparse sets of predictors are well known to perform better for high-dimensional problems \citep{guyon2003introduction}. In Section \ref{sec:variable_importance_selection} we experimentally show that our model performance is little affected by a large number of irrelevant variables.

A usual step after model selection is refitting of the parameters using the pooled training and validation data. In our model this is easy to perform. For each dimension we keep the learned functional structure fixed, i.e., we use exactly the same trees. We then simply rerun the updates of the leaf values according to equation (\ref{eq:leaf_update}). This refitting is computationally efficient because the most expensive part of training, namely deciding which split to perform, can be skipped. Updating the model parameters with additional data is less natural in neural networks. One heuristic to choose the number of iterations for the combined data is to learn until the average training likelihood from the earlier validated model is reached \citep{uria2014deep}. However, this does not really specify a particular model complexity and the approach will not work well if the validation data is easier or harder to fit than the training data. In our model such a retraining would correspond to learning new tree structures, which is a much slower process than simply updating the parameters.

Finally, boosted trees provide a simple way to quantify the importance of dependent variables. It is customary to sum up the gain (\ref{eq:gain}) for each predictor, which corresponds to the reduction in uncertainty due to that variable. In Section \ref{sec:variable_importance_selection} we exemplify the usefulness of this tool. We are not aware of a comparable feature importance measure for neural networks.

\section{Experiments}
\label{sec:experiments}

\subsection{Quantitative performance}
We evaluate our LogitBoost autoregressive network (LBARN) on 9 standard benchmark datasets using the same splits into training, validation and test sets as in previous work. The ratio of sample size and dimensionality in these datasets varies between 1 and 300, see Table \ref{tab:results} for details. The MNIST dataset \citep{salakhutdinov2008quantitative} was binarized through random sampling based on the original intensity values. The OCR-letters dataset comes from the Stanford AI Lab\footnote{\url{http://ai.stanford.edu/~btaskar/ocr/}}
and the remaining seven datasets come from the UCI Machine Learning Repository \citep{lichman2013uci}. We first focus on a quantitative comparison with neural autoregressive models. The standard way to assess the quality of probabilistic models is by evaluating the likelihood of the test data under the trained models. In principle, other proper scoring rules \citep{bickel2007some} could be considered but there are no previously reported results for alternative scoring rules. Reported likelihoods for the used datasets can be found in \citet{germain2015made} and \citet{bornschein2015reweighted}.

LBARN was trained using up to $T=1{,}000$ rounds of boosting per dimension. The number of trees to be used for each dimension was chosen through the model selection procedures presented in Section \ref{sec:model_selection} (we used the forward variant of the linearized selection procedure). For the number of leaves per tree we considered $J \in \{2,4,8,16,32,64,128\}$ (for balanced trees this corresponds to depths of 1-7) and selected the best value based on validation performance. The shrinkage factor was set to $\nu=0.02$. This value turned out to be small enough to train trees with a large number of leaves. Even smaller values of $\nu$ did not lead to better results but only slowed down the learning process. In cases where convergence had not occurred after 1,000 rounds of boosting we increased $\nu$ accordingly. This was only necessary for two datasets.

\afterpage{
\begin{landscape}
\begin{table*}[tb]
\caption{Average log-likelihoods (in nats) per test example for various models and datasets. The best result (without refitting) for each dataset is shown in bold. Dataset sizes, standard errors (for LBARN with individual model selection) and selected hyperparameters are shown in italic. $J$ is the selected number of leaves per tree and $\nu$ is the shrinkage factor.}
\label{tab:results}
\begin{center}
\begin{tabular}{l*{10}{c}}
\hline
& \bf Adult & \bf Connect4 & \bf DNA & \bf MNIST & \bf Mushrooms & \bf NIPS-0-12 & \bf OCR-letters & \bf RCV1 & \bf Web\\
\hline
\em train & \em 5,000 & \em 16,000 & \em 1,400 & \em 50,000 & \em 2,000 & \em 400 & \em 32,152 & \em 40,000 & \em 14,000\\
\em valid & \em 1,414 & \em 4,000 & \em 600 & \em 10,000 & \em 500 & \em 100 & \em 10,000 & \em 10,000 & \em 3,188\\
\em test & \em 26,147 & \em 47,557 & \em 1,186 & \em 10,000 & \em 5,624 & \em 1,240 & \em 10,000 & \em 150,000 & \em 32,561\\
\em dim & \em 123 & \em 126 & \em 180 & \em 784 & \em 112 & \em 500 & \em 128 & \em 150 & \em 300\\
~\\
Ber. Mix. & -20.44 & -23.41 & -98.19 & -137.64 & -14.46 & -290.02 & -40.56 & -47.59 & 30.16\\
FVSBN & -13.17 & -12.39 & -83.64 & -97.45 & -10.27 & -276.88 & -39.30 & -49.84 & -29.35\\
NADE & -13.19 & -11.99 & -84.81 & -88.33 & -9.81 & -273.08 & -27.22 & -46.66 & -28.39\\
EoNADE & -13.19 & -12.58 & -82.31 & -85.10 & -9.69 & -272.39 & -27.32 & -46.12 & -27.87\\
DARN & -13.19 & -11.91 & -81.04 & $\approx$\bf-84.13 & \bf -9.55 & -274.68 & $\approx$-28.17 & $\approx$-46.10 & $\approx$-28.83\\
MADE & -13.12 & -11.90 & -79.66 & -86.64 & -9.68 & -277.28 & -28.34 & -46.74 & -28.25\\
RWS-NADE & $\approx$-13.16 & $\approx$\bf-11.68 & $\approx$-84.26 & $\approx$-85.23 & $\approx$-9.71 & $\approx$\bf-271.11 & $\approx$-26.43 & $\approx$-46.09 & $\approx$-27.92\\
\hline
single tree & -13.32 & -13.83 & -80.72 & -93.41
 & -9.59 & -283.14 & -26.95 & -49.00 & -29.65
\\
LBARN\\
\hspace{1em}individual & -13.09 & -12.10 & -78.79 & -86.69 & -9.62 & -272.95 & \bf -24.60 & \bf -45.51 & \bf -27.60\\
\hspace{1em}common & -13.29 & -12.54 & \bf -78.64 & -87.29 & -9.71 & -271.53 & -24.74 & -45.56 & -27.67\\
\hspace{1em}linearized & \bf -13.07 & -12.14 & -78.91 & -86.86 & -9.71 & -271.72 & -24.63 & -45.54 & -27.65\\
LBARN refit & -13.07 & -12.06 & -77.93 & -86.49 & -9.54 & -272.32 & -24.47 & -45.35 & -27.40\\
~\\
\em std. error & \em 0.023 & \em 0.007 & \em 0.320 & \em 0.215 & \em 0.023 & \em 0.488 & \em 0.085 & \em 0.058 & \em0.103\\
\em J & \em 2 & \em 4 & \em 16 & \em 32 & \em 8 & \em 2 & \em 64 & \em 128 & \em 32\\
\em $\nu$ & \em 0.1 & \em 0.5 & \em 0.02 & \em 0.02 & \em 0.02 & \em 0.02 & \em 0.02 & \em 0.02 & \em 0.02
\end{tabular}
\end{center}
\end{table*}
\end{landscape}
}

\begin{figure*}[t]
\begin{center}
\includegraphics[height=.33\textheight]{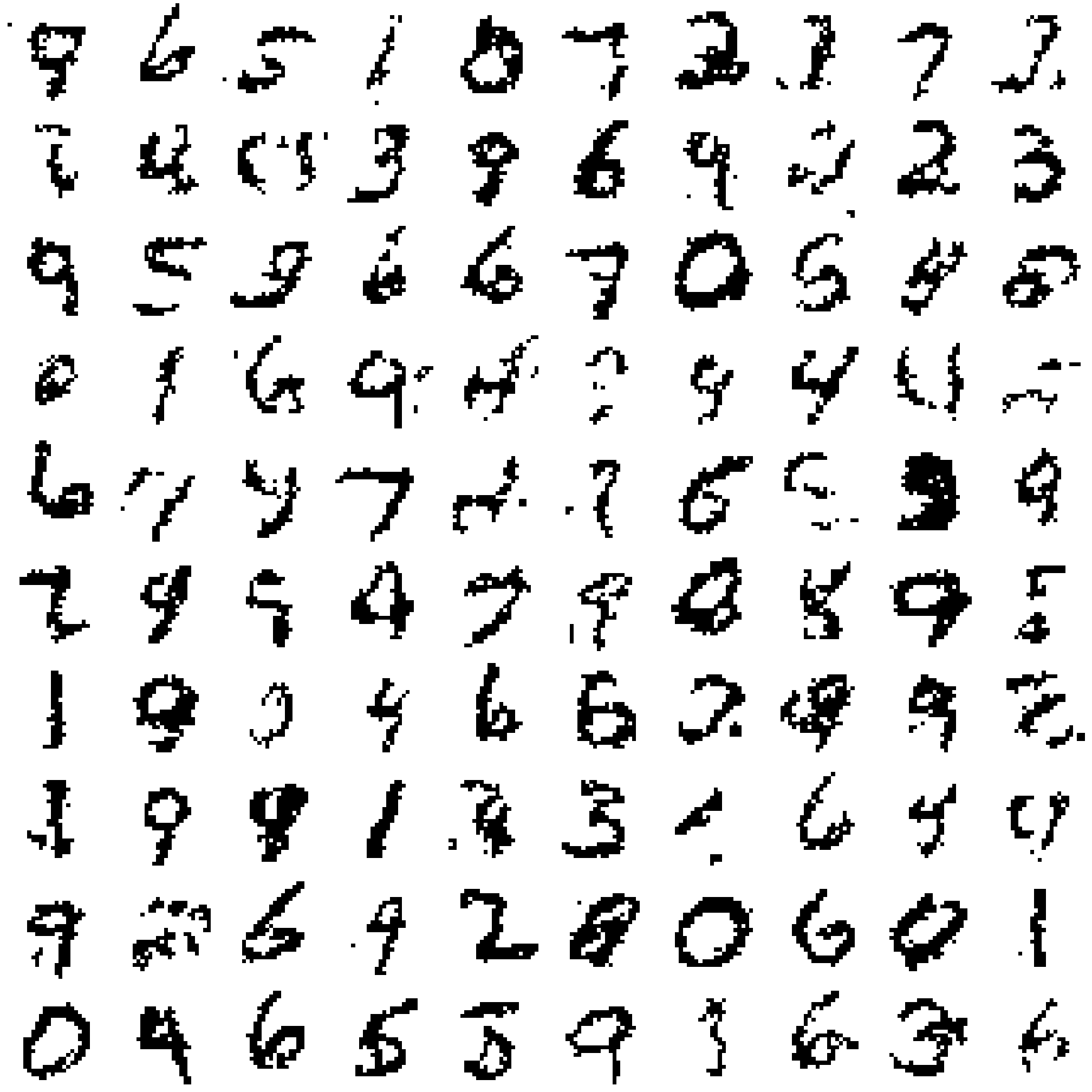}
\hfil
\includegraphics[height=.33\textheight]{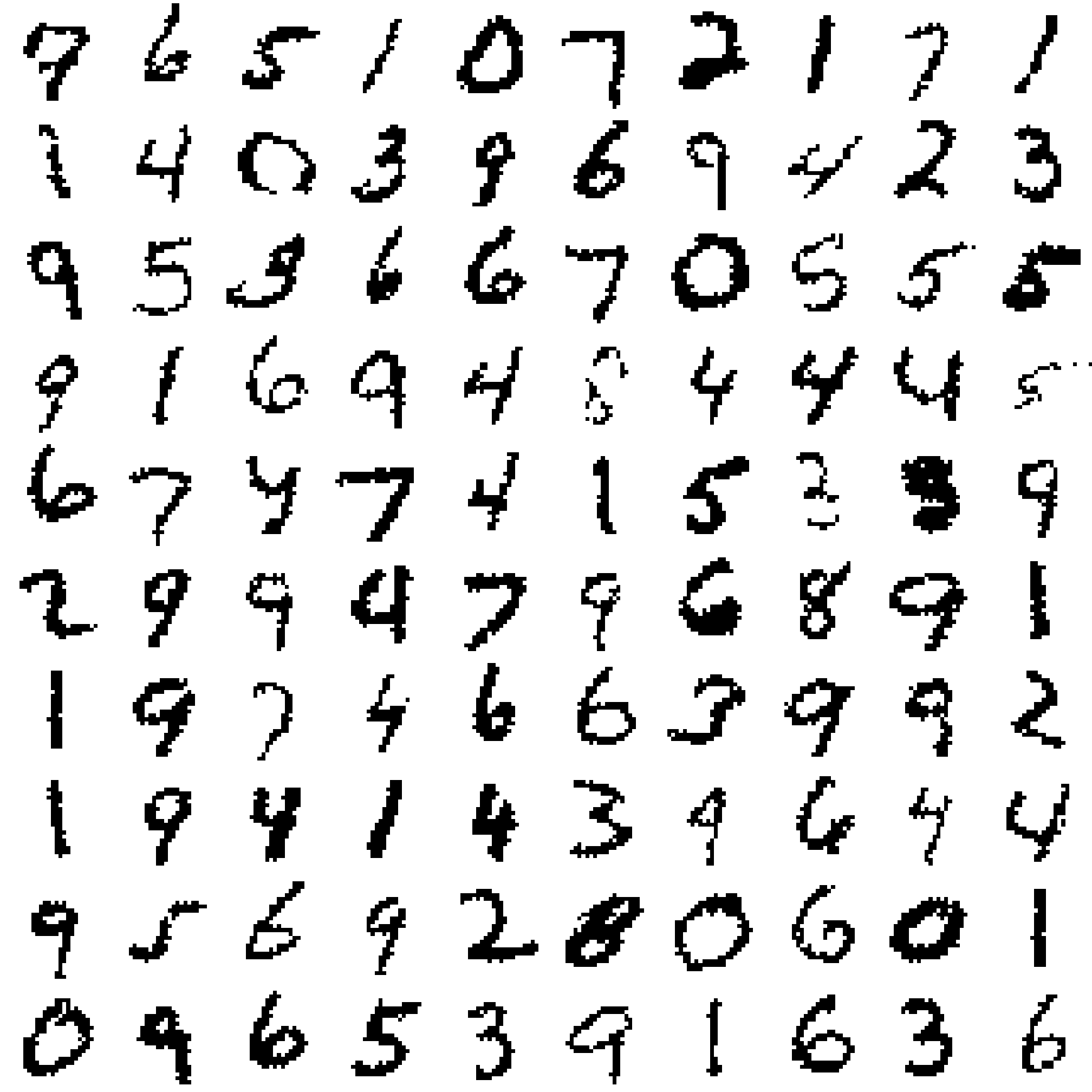}
\end{center}
\caption{\textbf{Left:} Samples from LBARN trained on MNIST digits. \textbf{Right:} Closest training examples.}
\label{fig:samples}
\end{figure*}

Table \ref{tab:results} reports our obtained test log-likelihoods together with previous results. We also included the selected number of leaves $J$ for each dataset and the used shrinkage factor $\nu$. For datasets with somewhat larger sample size (relative to the data dimension) deeper trees turned out to be beneficial. Table \ref{tab:results} also shows the test log-likelihoods after refitting of the parameters (in the network with individual model selection) using the pooled training and validation data. This always led to a small improvement. When comparing with previous results for fairness we only considered the model without refitting. Overall LBARN is among the best performing autoregressive models. On 5 datasets our results are better than the best reported result for neural autoregressive networks. On the other 4 datasets the performance is competitive with the state of the art. For the MNIST digits our model performs better than a neural network with a single layer of hidden units (NADE) and as good as a two-layer autoencoder with randomized connectivity pattern (MADE). The models that perform better than LBARN use a large ensemble of models for various orderings of the data dimensions (EoNADE) or rely on stochastic hidden units, which makes exact likelihood computations intractable (DARN, RWS-NADE). In Figure \ref{fig:samples} we show samples of the learned model for MNIST digits, together with nearest neighbors from the training set.

As an additional comparison we trained autoregressive networks using a single probability estimation tree \citep{provost2003tree} for each dimension. The number of leaves for each tree was determined based on the performance on the validation set. In addition, small pseudocounts were used to regularize the probabilities. For each dataset the same regularization strength was chosen for all trees using the validation set. The obtained test log-likelihoods of these single-tree autoregressive networks can also be found in Table \ref{tab:results}. With one exception boosted trees performed much better than single trees.

The performance of the three model selection procedures is overall comparable. Individual validation for each dimension worked well in most cases. Overfitting of the validation set only occurred for the NIPS-0-12 dataset, which has 500 dimensions but contains merely 100 validation samples. Selection of a common number of trees for all dimensions was often the worst option. The linearized model selection worked almost as well as individual selection and performed better for datasets with a small number of validation samples. The standard errors reported in Table \ref{tab:results} for the test log-likelihoods of LBARN correspond to the network with individual model selection.

\begin{figure}[t]
\begin{center}
\includegraphics[height=.25\textheight]{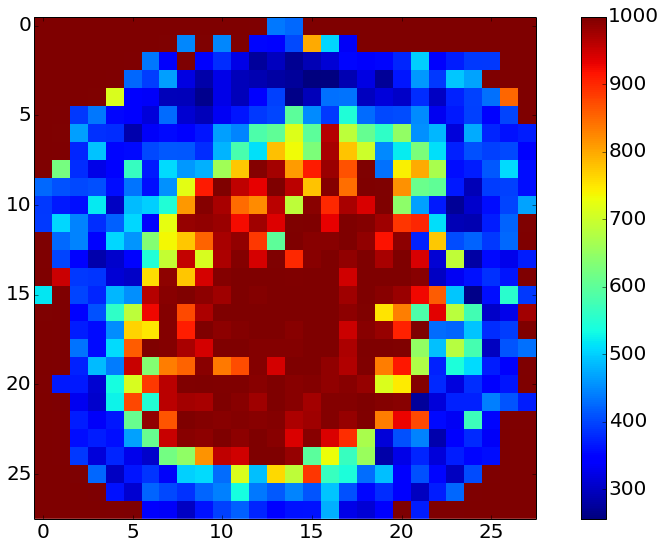}
\end{center}
\caption{Map of the selected numbers of trees for the MNIST dataset.}
\label{fig:ntrees}
\end{figure}

Figure \ref{fig:ntrees} shows a map of the selected number of trees at each pixel for the MNIST dataset. For many dimensions around 400 trees were enough. However, a large fraction of the conditionals use close to 1,000 trees, which is the total number of boosting rounds. In those cases the holdout performance saturated. That means more trees do not lead to overfitting but the additional gain is very small. To confirm this experimentally, we continued training for up to 2,000 rounds of boosting. The test log-likelihood increased slightly from the original -86.69 to -86.62.

\begin{figure}
\begin{center}
\includegraphics[height=.25\textheight]{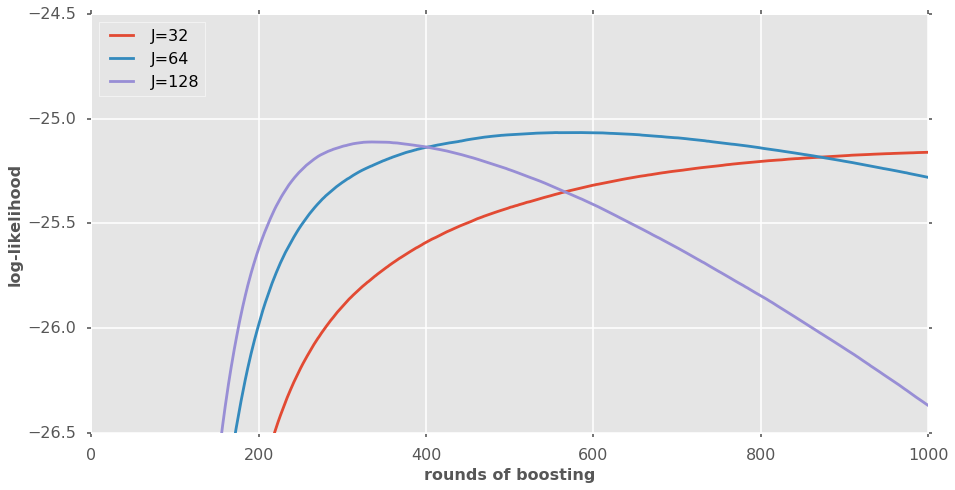}
\hfil
\includegraphics[height=.25\textheight]{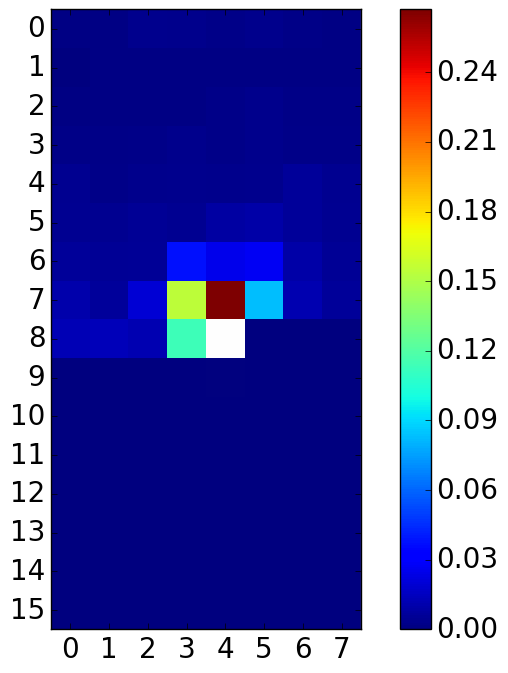}
\end{center}
\caption{\textbf{Left:} Validation log-likelihood on the OCR-letters dataset for different numbers of leaves. \textbf{Right:} Variable importance for the central pixel (white) of the OCR-letters data.}
\label{fig:ocr}
\end{figure}

The main hyperparameter, which we have to chose for LBARN, is the number of leaves $J$ for each tree. In the left panel of Figure \ref{fig:ocr} we show the validation log-likelihood on the OCR-letters dataset as a function of boosting rounds for three different choices of $J$. The shrinkage factor was always $\nu=0.02$. For $J=32$ LogitBoost did not fully converge after 1,000 rounds of boosting. For $J=64$ the best validation performance is obtained with around 600 rounds of boosting, after that slight overfitting begins. For $J=128$ strong overfitting occurs after 400 rounds of boosting. Importantly, the maximum of all three curves is almost the same. So, (at least in this example) the exact value of $J$ is not that important as long as it is in the appropriate range.

A variant of LogitBoost \citep{friedman2000additive} is Gradient Boosting \citep{friedman2001greedy}, which approximates the gradient of the Bernoulli log-likelihood to learn boosted trees rather than using a second-order expansion. We ran a few experiments with this alternative boosting method. The performance was usually a bit worse than with LogitBoost. This confirms the observation made by \citet{li2010robust} that second-order information improves the results.

\subsection{Variable importance and variable selection}
\label{sec:variable_importance_selection}
The splitting gain (\ref{eq:gain}) is a useful tool to quantify variable importance. We illustrate this through the model trained for the OCR-letters data (consisting of images of size $16 \times 8$). In the autoregressive network the central pixel (located at row 8, column 4) is modeled conditional on all pixels that are earlier in the rowwise ordering (i.e. that are above or to the left). As shown in the right panel of Figure \ref{fig:ocr}, the dependencies are mostly local. Indeed, the four nearest neighbors (left, top-left, top and top-right pixel) together explain about 62\% of the uncertainty. Being able to quantify the dependencies between the variables is a useful feature, which improves the understanding of the dataset.

We performed an additional experiment to illustrate that the conditional distributions in LBARN can be learned robustly even when a large number of irrelevant variables is present. The OCR-letters data has 128 dimensions. By stacking three independent copies next to each other we created a new dataset with 384 dimensions, corresponding to the product of three (independent) OCR-letters distributions. Specifically, the training, validation and test datasets are formed by using the original data for dimensions 1-128 and by using two different shufflings of the sample orders for dimensions 129-256 and 257-384, respectively. The best achievable log-likelihood for this dataset is three times the log-likelihood of the OCR-letters data. If during the training phase spurious relationships between the copies are found then the training log-likelihood will be larger than three times the training log-likelihood of the OCR-letters data. However, in that case the test likelihood will be smaller than three times the test log-likelihood of the OCR-letters data. For the first copy $\bm{x}_{1:128}$ the average test log-likelihood we achieved with our trained model was -24.60, as before. In our autoregressive network, the second copy $\bm{x}_{129:256}$ is estimated conditional on the first 128 (irrelevant) variables and the third copy $\bm{x}_{257:384}$ is estimated conditional on 256 nuisance variables. The test log-likelihood for the second (-25.25) and third (-25.42) copy are only a bit lower compared to the first copy. In particular, the performance is still better than the best result from the alternative models in Table \ref{tab:results}. A possible reason why the performance decreased at all is that all our trees have the same number of leaves. Consequently, some splitting variables have to be chosen. By requiring a minimum amount of improvement for each additional leaf we believe that the results in the presence of irrelevant variable can be further improved.

\subsection{Ordering of the variables}

\begin{figure*}[t]
\begin{center}
\includegraphics[height=.25\textheight]{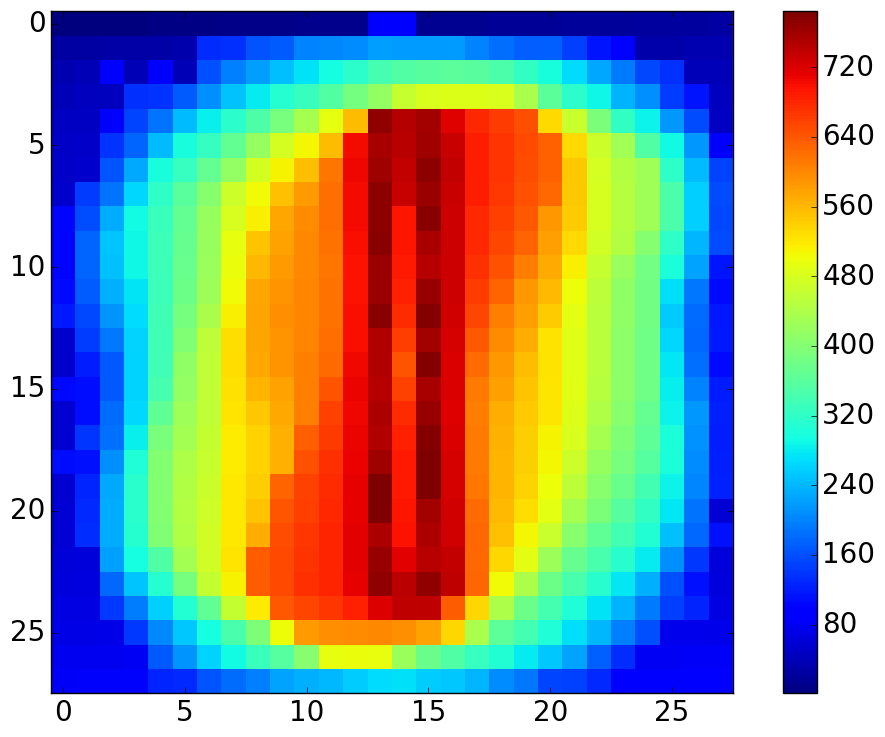}
\hfil
\includegraphics[height=.25\textheight]{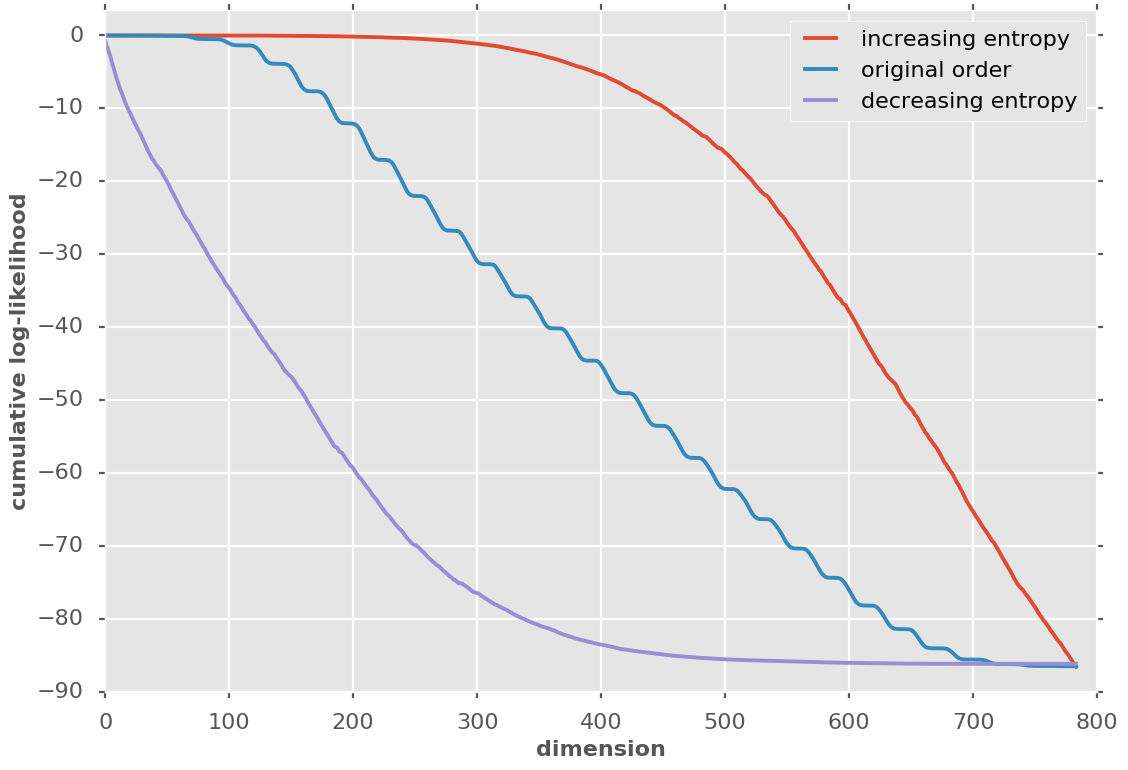}\\
\includegraphics[height=.25\textheight]{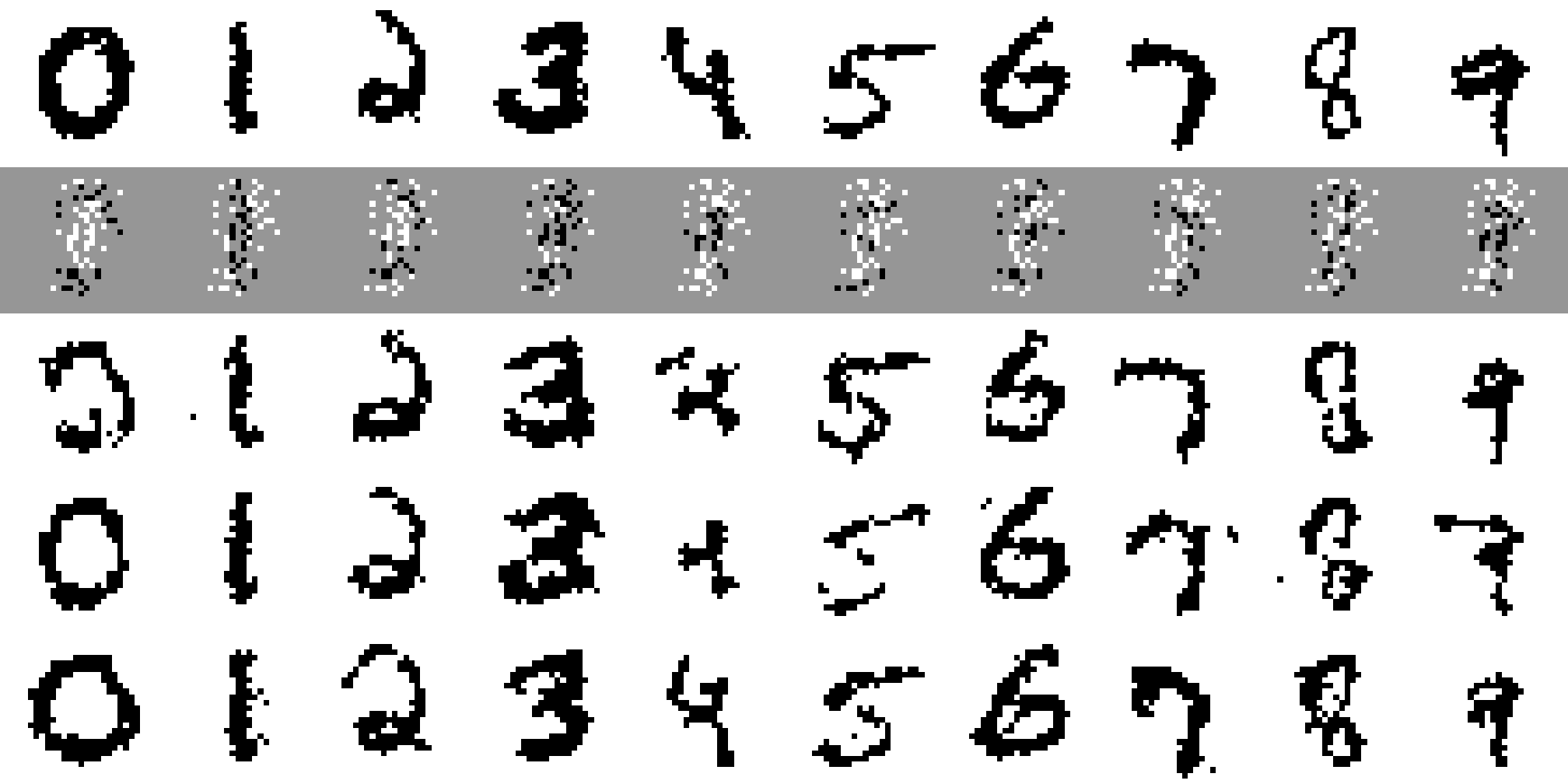}
\end{center}
\caption{\textbf{Top-left:} Pixels for MNIST digits ordered by conditional entropy (blue=low entropy, red=large entropy). \textbf{Top-right:} Cumulative log-likelihoods of the MNIST test digits for different pixel orderings. \textbf{Bottom:} Compression of MNIST digits. 1st row: Test examples. 2nd row: Reduced data consisting of 78 pixels (gray indicates missing values). 3rd-5th rows: Generated samples conditioned on the reduced data.}
\label{fig:mnist_ordering}
\end{figure*}

For any permutation $\pi$ of the dimensions $\{1,\ldots,D\}$ the product of conditional distributions
\[
\prod_{d=1}^D \mathbb{P}(x_{\pi(d)} \,|\, x_{\pi(1)},\ldots,x_{\pi(d-1)})
\]
is equal to the joint distribution $\mathbb{P}(\bm{x})$. However, if the conditionals are learned from data then different decompositions can yield different results. So far we always used the ordering of the variables in which they are stored in the computer. For specific applications alternative orderings can be more appropriate. For example, by using an ordering in which the first few variables provide as much information as possible about the complete data we can create a simple (noisy) compression mechanism. Such an ordering is characterized by the fact that the later conditional distributions have a low entropy. A greedy strategy to find an ordering with that property is to sequentially add the variable with the highest conditional entropy given all previously added variables \citep{geman1996active}.

We used the training examples from the MNIST digits dataset to sort the $28 \times 28 = 784$ variables by conditional entropy. The top-left panel of Figure \ref{fig:mnist_ordering} visualizes the obtained ordering. We consider the increasing-entropy order, which starts from background pixels at the boundary of the image grid and circulates towards the inside, as well as the decreasing-entropy order, which starts from central pixels and moves outwards. To illustrate the captured amount of information we plot in the top-right panel of Figure \ref{fig:mnist_ordering} the cumulative log-likelihoods $\sum_{d=1}^D \log\mathbb{P}(x_{\pi(d)} \,|\, x_{\pi(1)},\ldots,x_{\pi(d{-}1)})$ as a function of $D$. When sorted by increasing conditional entropy the variables at the end of the ordering contribute most. When sorted by decreasing conditional entropy most of the contributions to the log-likelihood come from variables at the beginning of the ordering. Nevertheless, the average test log-likelihoods for the full data are rather similar. The original ordering leads to a test log-likelihood (after refitting) of -86.49. With the increasing-entropy order the log-likelihood is slightly lower (-86.53) while with the decreasing-entropy order it is 1.6 standard errors higher (-86.15). The advantage of having the most informative variables early in the ordering is that we can reduce the data to a small fraction of the dimensions and can still reconstruct the full data well. This is illustrated in the bottom panel of Figure \ref{fig:mnist_ordering}, where we use examples from the MNIST test set and generate samples conditioned on the first 10\% of the variables.

\section{Discussion}
We introduced a novel autoregressive network that achieves performance comparable to or even better than the best existing models. Remarkably, this was possible by using separate conditional distribution estimators with no weight sharing. The choice of LogitBoost for modeling the conditionals turned out to be fruitful. However, we conjecture that competitive results can be obtained with other state-of-the-art probability estimators. The specific advantages of our approach compared to neural autoregressive models are simplicity, interpretability and scalability. The Python implementation of LBARN is available at \url{http://github.com/goessling/lbarn}.

We believe that our model can be further improved by using recently developed boosting extensions like Bayesian additive regression trees \citep{chipman2010bart}, dropouts for additive regression trees \citep{rashmi2015dart} or feature subsampling for LogitBoost \citep{chen2016xgboost}. Moreover, rather than starting the boosting process with log-odds of 0 we could initialize the model by first fitting a logistic regression of $x_d$ on $\bm{x}_{1:d}$. The obtained predicted probabilities can then be used as the base probabilities for boosting. Such an initialization potentially decreases the required number of boosting rounds. It also allows for linear effects of the predictors instead of piecewise constant relations. In neural networks this roughly corresponds to introducing direct connections between $\bm{x}_{1:d-1}$ and $x_d$.

Weight sharing between conditional distributions for different dimensions or weight sharing between models trained for different orderings of the variables \citep{uria2014deep} is orthogonal to our approach. However, by replacing the decision trees in our model with directed acyclic graphs \citep{busa2012fast,shotton2013decision} it is in principle possible to introduce weight sharing into our network.


In neural autoregressive networks for real-valued data \citep{uria2013rnade} weight sharing among conditional distributions has been shown to be essential for regularization. The natural extension of our approach to real-valued data is to train separate regression estimators for each dimension. This could, for example, be done using L2-boosting \citep{friedman2001greedy,buhlmann2003boosting}, Gaussian process regression \citep{rasmussen2006gaussian} or sparse additive models \citep{ravikumar2009sparse}. Note that in addition to the conditional mean one also has to estimate the conditional variance, assuming that a simple parametric model for the conditional distributions is used. Alternatively, a multi-class LogitBoost method could be used if the real values are discretized.

\section*{Acknowledgements}
I am grateful to Yali Amit for his comments on this work. I also thank the anonymous reviewers for their valuable feedback, which helped to improve the manuscript.

\section*{References}
{\small
\bibliography{references}}

\end{document}